\documentclass{article}

\usepackage{microtype}
\usepackage{graphicx}
\usepackage{subfigure}
\usepackage{amsmath}
\usepackage{subcaption}
\usepackage{stfloats} 
\usepackage{xcolor}
\usepackage{booktabs} 

\usepackage{hyperref}

\usepackage[accepted]{icml2024}

\usepackage{amsmath}
\usepackage{amssymb}
\usepackage{mathtools}
\usepackage{amsthm}

\usepackage[capitalize,noabbrev]{cleveref}

\theoremstyle{plain}

\theoremstyle{definition}

\theoremstyle{remark}

\def\enc{\mathbf{u}_\textrm{enc}}
\def\dec{\mathbf{u}_\textrm{dec}}

\def\nprcc{\texttt{NeRCC}}
\def\func{\mathbf{f}}
\def\fhat{\mathbf{\hat{f}}}

\usepackage[textsize=tiny]{todonotes}

\icmltitlerunning{{Nested-Regression Coded Computing for Resilient Distributed Prediction Serving Systems}}

\begin{document}
\twocolumn[
\icmltitle{{\LARGE \texttt{NeRCC}}: Nested-Regression Coded Computing for Resilient Distributed Prediction Serving Systems}




\icmlsetsymbol{equal}{*}

\begin{icmlauthorlist}
\icmlauthor{Parsa Moradi}{yyy}
\icmlauthor{Mohammad Ali Maddah-Ali}{yyy}
\end{icmlauthorlist}

\icmlaffiliation{yyy}{Department of Electrical and Computer Engineering, University of Minnesota, Minneapolis, USA}

\icmlcorrespondingauthor{Mohammad Ali Maddah-Ali}{maddah@umn.edu}

\icmlkeywords{Distributed Computing, Machine Learning, ICML}

\vskip 0.3in
]



\printAffiliationsAndNotice{}  

\begin{abstract}
Resilience against stragglers is a critical element of prediction serving systems, tasked with executing inferences on input data for a pre-trained machine-learning model. In this paper, we propose $\nprcc$, as a general straggler-resistant framework for approximate coded computing. $\nprcc$ includes three layers: (1) encoding regression and sampling, which generates coded data points, as a combination of original data points, (2) computing, in which a cluster of workers run inference on the coded data points, (3) decoding regression and sampling, which approximately recovers the predictions of the original data points from the available predictions on the coded data points. 
We argue that the overall objective of the framework reveals an underlying interconnection between two regression models in the encoding and decoding layers. We propose a solution to the nested regressions problem by summarizing their dependence on two regularization terms that are jointly optimized. Our extensive experiments on different datasets and various machine learning models, including LeNet5,  RepVGG, and Vision Transformer (ViT), demonstrate that $\nprcc$ accurately approximates the original predictions in a wide range of stragglers, outperforming the state-of-the-art by up to 23\%. 
\end{abstract}

\section{Introduction}
Prediction serving systems are infrastructures, designed to deliver the predictions of pre-trained machine learning models for new input data points. 
These systems ensure efficiency and scalability by distributing input data across multiple workers, each running an instance of the model. One of the primary challenges raised in these distributed computing systems is the presence of slow-performing workers, commonly referred to as stragglers. In such instances, the prediction serving system must wait for all workers to respond, thereby impacting the overall latency of the process.

To address this issue, these systems often employ the strategy of replicating input data points, sending the same data points to multiple workers \cite{suresh2015c3, shah2015redundant}. However, this replication introduces significant overhead and inefficiency to the system. Recently, coded computing methods have been introduced that significantly reduce system overhead and offer efficient resilience to stragglers~\cite{lee2017speeding, kosaian2018learning, narra2019distributed, yu2019lagrange}. However, these methods have been originally limited to exact matrix multiplication and polynomial computation and require a large number of workers. Another line of research has extended coded computing to more general computations, allowing for the relaxation of result precision to approximation~\cite{kosaian2019parity,so2020scalable, jahani2022berrut}. This approach is utilized to reduce the overhead and latency of prediction-serving systems~\cite{kosaian2019parity, dinh2021coded, jahani2022berrut, soleymani2022approxifer}. In this approach, a master node in the system computes a set of coded data points, each as a combination of all input data. Subsequently, each coded data is sent to individual workers, where the worker runs the model on the assigned coded data point and forwards the predictions back to the master node. The coding scheme is designed to ensure that even if some workers are stragglers, the master node can still recover (or decode) the approximation of the model's predictions for the original data with relatively small errors. In \cite{kosaian2019parity}, \texttt{ParM}, as a model-specific approach, is introduced, wherein a parity model is trained on coded data to precisely predict the coded output, enhancing straggler tolerance. However, this approach faces scalability challenges, as a new parity model needs to be trained for each new model. Furthermore, it only tolerates one single straggler, and the experimental results are reported for a small number of workers. Inspired by rational interpolation \cite{berrut1988rational}, \cite{jahani2022berrut} proposed a model-agnostic and numerically stable framework known as \texttt{BACC} for general computations, and successfully employed it to train neural networks on a cluster of workers, while tolerating a larger number of stragglers. Building on \texttt{BACC} framework, \cite{soleymani2022approxifer} 
introduced \texttt{ApproxIFER} scheme, as a straggler resistance and Byzantine-robust prediction serving system. \texttt{BACC} framework is designed based on numerically stable \emph{exact} interpolation, using pole-free rational interpolations. While the \texttt{BACC} framework is designed for approximate computing, it does not prioritize the accuracy of the prediction for the input data points; instead, it places unnecessary emphasis on the exactness of the interpolations.

In this paper, we introduce \emph{Nested-Regression Coded Computing} (\nprcc), a general and straggler-resistant framework for approximate distributed coded computing, employed in prediction serving systems. This framework consists of three layers: (1) encoding regression and sampling, generating coded data points by running a regression on input data points, and sampling from the resulting regression function. (2) Computing, where each worker runs the trained machine learning model on one of the coded data points, (3) decoding regression and sampling, generating the approximate prediction by running a regression on available predictions on the coded data points. We argue that the estimation error of the prediction system can be upper-bounded by the summation of two interconnected terms: one reflecting the training error of the encoding regression and the other representing the generalization error of the decoding regression. The coupling between these nested regressions complicates their design and analysis. To take this coupling into account, and still leverage the established science of regression models,  we formulate the problem as two regression tasks, summarizing their interdependence
in two regularization terms. However, the weight of the regularizing terms is optimized jointly. 

We extensively evaluate the performance of $\nprcc$ for a wide range of straggler numbers and various machine learning models. These models include LeNet5~\cite{lecun1998gradient}, with 1024-dimensional inputs and 10-dimensional outputs, having $6\times10^4$ parameters; RepVGG~\cite{ding2021repvgg} with 3072-dimensional inputs, 10-dimensional outputs, and 26 million parameters; and Vision Transformer (ViT)~\cite{dosovitskiy2020image} with 150528-dimensional inputs, 1000-dimensional outputs, and 80 million parameters. Our results demonstrate that the proposed framework consistently outperforms the state of the art in all different scenarios. This improvement is reflected in enhanced accuracy and a notable reduction in the mean square error of approximate models, achieving up to 6\% and 23\% respectively, compared to the Berrut interpolation approaches.

\begin{figure*}[ht]
\vskip 0.2in
\begin{center}
    \centerline{\includegraphics[width=0.6\textwidth]{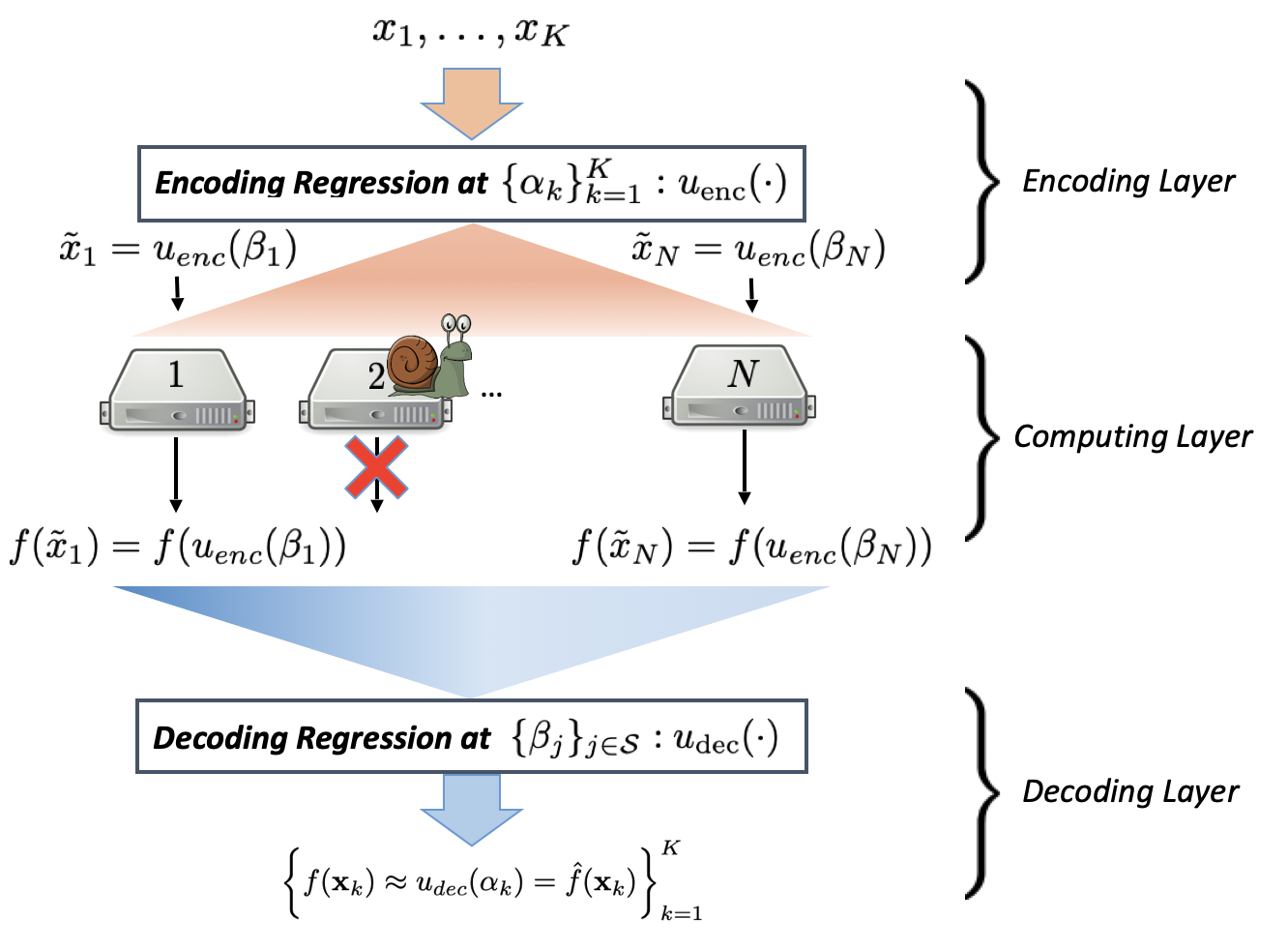}}
    \caption{$\nprcc$ framework. It consists of a computing layer sandwiched between two regression layers.}
    \label{fig:framework}
\end{center}
\vskip -0.3in
\end{figure*}

\section{Method}
\subsection{Notations}
Throughout this paper, uppercase and lowercase bold letters denote matrices and vectors, respectively. Coded vectors and matrices are indicated by a $\sim$ sign, as in $\tilde{\mathbf{x}}, \tilde{\mathbf{A}}$. The $i$-th element of a vector $\mathbf{x}$ is denoted as $x_i$, and the $(i_1, i_2, \dots, i_d)$-th element of a $d$-dimensional matrix $\mathbf{A}$ is represented by $A_{i_1, i_2, \dots, i_d}$. Additionally, $x^i_j$ represents the $j$-th element of the $i$-th vector in the set $\{\mathbf{x_k}\}^n_{k=1}$. The set $\{1, 2, \dots, n\}$ is denoted as $[n]$. The symbol $|S|$ denotes the cardinality of the set $S$.
The $k$-th derivative function of the one-dimensional function $f(\cdot)$ is denoted by $f^{(k)}(\cdot)$.

\subsection{Problem Setting}
Consider a master node and a set of $N$ workers. The master node is provided with a set of $K$ data points $\{\mathbf{x}_k\}_{k=1}^K \in \mathbb{R}^d$ to compute $\{\func(\mathbf{x}_k)\}^K_{k=1}$ using a cluster of $N$ worker nodes. Here, $\func: \mathbb{R}^d \to \mathbb{R}^m$ represents an arbitrary function, and $K$, $d$, and $m$ are integers. In prediction serving systems, the function $\func(\mathbf{x})$ represents a pre-trained machine learning model, for example, a deep neural network model, where $\func(\mathbf{x})$ is the inference process for a given input $\mathbf{x}$. A naive approach is to assign these computations to the workers to execute them. The challenge is that some of the worker nodes may act as stragglers and not finish their tasks within the required deadline. A promising approach for the master node to mitigate this problem is to use coding in its computation. In this approach, the master node offloads some (linear) combination of the initial data points, rather than raw data points, to the worker nodes, collects their prediction results, and then decodes the predictions of the original data from the predictions on the (linear) combinations. The goal of the master node is to approximately compute $\hat{f}(\mathbf{x}_k) \approx \func(\mathbf{x}_k)$ at the data points $\{\mathbf{x}_k\}_{k=1}^K$, even if some of the worker nodes appear to be stragglers.

\subsubsection{Proposed Framework}\label{sec:framework}
Here we propose a general straggler-resistant framework for distributed coded computing. This framework has two layers of regressions and a computing layer, where the computing layer is sandwiched between two regression layers. This framework is motivated by \texttt{BACC} scheme~\cite{jahani2022berrut}, and includes it as a special case.  
This scheme is based on the following steps:
\begin{enumerate}

    \item \textbf{Encoding regression:} The master node fits a regression function $\enc: \mathbb{R} \to \mathbb{R}^d$ at points $\{(\alpha_k, \mathbf{x}_k)\}^K_{k=1}$ for distinct ordered values $\alpha_1 < \alpha_2 < \dots < \alpha_{K} \in \mathbb{R}$. Then, it computes the encoding function $\enc(\cdot)$ on a set of distinct ordered values $\{\beta_n\}_{n=1}^N$ where $\beta_1 < \beta_2 < \dots < \beta_{N} \in \mathbb{R}$, with $k\in[K]$ and $n\in[N]$. Subsequently, the master node sends $\tilde{\mathbf{x}}_n = \enc(\beta_n) \in \mathbb{R}^d$ to worker $n$. It is noteworthy that each coded point $\tilde{\mathbf{x}}_n$ is a linear combination of all initial points $\{\mathbf{x}_k\}^K_{k=1}$.
    
    \item \textbf{Computing at worker nodes:} Each worker $n$ computes $\func(\Tilde{\mathbf{x}}_n) = \func(\enc(\beta_n))$ on its given input for $n \in [N]$ and returns the result to the master node.
    
    \item \textbf{Decoding regression:} The master node receives the results ${\func(\tilde{\mathbf{x}}_j)}_{j\in \mathcal{S}}$, where $\mathcal{S}$ represents a set of non-straggler worker nodes. Next, it fits a decoding regression function $\dec: \mathbb{R} \to \mathbb{R}^m$ at points ${(\beta_{s}, \func(\tilde{\mathbf{x}}_j))}_{j\in \mathcal{S}}= {(\beta_{s}, \func(\enc(\beta_{j})))}_{j\in \mathcal{S}}$. Finally, using the function $\dec(\cdot)$, the master node can compute $\hat{f}(\mathbf{x}_k):=\dec(\alpha_k)$ as an approximation of $\func(\mathbf{x}_k)$ for $k \in [K]$. Recall that $\dec(\alpha_k) \approx \func(\enc(\alpha_k)) \approx \func(\mathbf{x}_k)$.
\end{enumerate}

The effectiveness of approximate coded computing frameworks in addressing the straggler issue relies on the fact that each worker receives a coded point, which is a combination of all original data points. With a well-designed framework, we anticipate that the master node will be capable of recovering an approximation of the function output at the initial points from the prediction results of non-straggler workers.

Additionally, it is crucial point that the encoding and decoding regressions must be efficient. This is because the overall computation on the master node, including developing the encoding and decoding functions, and calculating those regression functions at points $\{\beta_j\}_{j\in \mathcal{S}}$ and $\{\alpha_k\}^K_{k=1}$, 
must be less than the scenario where the master node computes $\{\func(\mathbf{x}_k)\}^K_{k=1}$ all by itself. Otherwise, adopting this framework would not yield benefits for the master node.

This general framework is motivated by Lagrange~\cite{yu2019lagrange} and Berrut~\cite{jahani2022berrut} coded computing approaches and encompasses them as a special case. Lagrange-coded computing~\cite{yu2019lagrange} is designed for exact polynomial computation, limited to polynomial encoding and decoding functions. In that scheme, the number of worker nodes needed grows with the degree of the polynomial multiplied by the number of data points, posing a scalability challenge for the entire scheme. These methods use fixed $\enc(\cdot)$ and $\dec(\cdot)$ as Lagrange polynomial interpolation functions. On the other hand, Berrut-coded computing approaches~\cite{jahani2022berrut} employ rational interpolation functions with desired numerical properties for $\enc(\cdot)$ and $\dec(\cdot)$. They can recover highly accurate approximations of ${\func(\mathbf{x}_k)}^K_{k=1}$ for an arbitrary function $\func(\cdot)$ even in the presence of straggler workers. However, in the Berrut interpolation framework, the coupling of encoding and decoding regressions, as well as the end-to-end objective, is not considered.

\subsection{Objective}\label{sec:objective}
As depicted in Figure~\ref{fig:framework}, the $\nprcc$ framework consists of a computing layer where an arbitrary function $\func(\cdot)$ is applied to the coded points $\{\tilde{\mathbf{x}}_n\}^N_{n=1} = \{\enc({\beta_n})\}^N_{n=1}$. This layer is sandwiched between two regression models. The overall performance of the scheme is measured by the following objective function:
\begin{align}\label{eq:main_formula}
    L = \sum^K_{k=1} \ell(\fhat(\mathbf{x}_k), \func(\mathbf{x}_k))=
    \sum^K_{k=1} \ell(\dec(\alpha_k), \func(\mathbf{x}_k)),
\end{align}
where $\ell: \mathbb{R}^m \times \mathbb{R}^m \to \mathbb{R}^+$ is a distance function that measures the distance between a pair of points. Utilizing the triangle inequality, an upper bound for the objective function~\eqref{eq:main_formula} can be derived:
\begin{align}
\nonumber
    L &\le \sum^K_{k=1} \underbrace{\ell\bigl(\dec(\alpha_k), \func(\enc(\alpha_k))\bigr)}_{\text{(Term 1)}} \\ \label{eq:decompose} &+ 
    \sum^K_{k=1} \underbrace{\ell\bigl(\func(\enc(\alpha_k)), \func(\mathbf{x}_k)\bigr)}_{\text{(Term 2)}}.
\end{align}

The right-hand side of the inequality \eqref{eq:decompose} reveals the underlying interplay between the encoding and decoding regression models of $\nprcc$ scheme, which is represented by two terms. The first term represents
\emph{the generalization error of decoding regression} and the second term acts as a proxy for \emph{the training error of encoding regression}. In what follows, we elaborate on each term: 
\begin{itemize}
    \item \textbf{Term 1 -- The generalization error of the decoding regression:} Recall that the master node fits a regression model at a set of points denoted as $\{(\beta_j, \func(\enc(\beta_j)))\}_{j \in \mathcal{S}}$. The first term on the right-hand side of \eqref{eq:decompose} represents the computation loss of the decoding regression model on points $\{\alpha_k\}^K_{k=1}$, which are distinct from $\{\beta_j\}_{j\in \mathcal{S}}$. Thus, it corresponds to the \emph{generalization error} of the regression model. 

Let us review two results on the generalization error for non-parametric regression. 

Consider a function $g : [0, 1] \to \mathbb{R}$, where $g(\cdot)\in \Sigma(\gamma, J)$, for an integer $\gamma \ge 1$ and a positive number $J$. The set $\Sigma(\gamma, J)$, known as the Hölder class, denotes the set of functions that are $\gamma - 1$ times differentiable, and the $\gamma - 1$ derivative of $g(\cdot)$ is $J$-Lipschitz continuous. Now, consider a regression model that constructs the regression function $\hat{g}(\cdot)$ using $n$ noisy samples of a function $g(\cdot)$. In \citep[p. 50]{alma9925110420001701}, it is demonstrated that for a local polynomial estimator with bandwidth $\delta$ representing the smoothness of the regression function, for every $x_0 \in [0, 1]$, the generalization error is upper-bounded as:
\begin{align}\label{eq:lp_error}     
    \mathbb{E}\bigl(||g(x_0)-\hat{g}(x_0)||_2^2\bigr) \le J^2C_1\delta^{2\gamma} + \frac{C_2}{n\delta},
\end{align}
 where $C_1$ and $C_2$ are two constants (for more details, refer to \citep[p. 50]{alma9925110420001701}). The inequality \eqref{eq:lp_error} illustrates how the smoothness of the function $g(\cdot)$ (characterized by $J$) and the smoothness of the regression function (indicated by the bandwidth parameter $\delta$) impact the generalization error.

Let us assume that     $g(\cdot) \in \mathbb{W}^s_2[a,b]$,  containing all functions for which the set $\{g^{(i)}(\cdot)\}_{i=0}^{s-1}$ is absolutely continuous and $\int_{a}^{b} (g^{(s)}(t))^2 \,dt < \infty$.
   
Then, \citep[p. 14]{ragozin1983error} established if we have $n$ samples of $g$ and use 
    $r$-th order smoothing spline as a regression model, equipped with a smoothing parameter $\lambda$, to derive $\hat{g}(\cdot)$ as estimation of $g(\cdot)$, then  for any $x_0 \in [a, b]$, we have  \begin{align}\label{eq:spline_error}
        ||g(x_0)-\hat{g}(x_0)||_2^2 \le &Q\bigl(\{x_i\}^n_{i=1}, \lambda, r, m, a, b\bigr) \nonumber  \\ 
        & \times  \int_{a}^{b} (g^{(s)}(t))^2 \,dt.
    \end{align}
To see the detailed formula of $Q(\cdot)$, refer to \citep[p. 14]{ragozin1983error}. 
Again, the above formula implies that the generalization error is a function of the smoothness of function $g(\cdot)$, represented by  $\int_{a}^{b} (g^{(s)}(t))^2 \,dt$ and smoothness of the regression function, represented by $\lambda$.

{\bf Observation One:} Since, in the decoding regression, the model is developed to estimate $\func(\enc(\cdot))$, the generalization error of the decoding regression depends on the derivatives of $\func(\enc(\cdot))$, which itself relies on the derivatives of both $\enc(\cdot)$ and $\func(\cdot)$, following the chain rule. As a result, the two terms in \eqref{eq:decompose} are inherently interconnected.

    \item \textbf{Term 2 - A proxy to the training error of the encoding regression:} 
    For simplicity of exposition, let's consider the mean square error (MSE) as the loss function $\ell(.,.)$. Using Taylor expansion, we can rewrite the second term as: 
\begin{align}\label{eq:taylor}  \sum_{k=1}^K||&\func(\enc(\alpha_k)) - \func(\mathbf{x}_k)||^2_2  \nonumber \\  & \approx \sum_{k=1}^K || (\enc(\alpha_k) - \mathbf{x}_k)^T\nabla \func(\mathbf{x}_k)||^2_2 \nonumber  \\ &   \le || \nabla \func||^2_{\infty}\sum_{k=1}^K ||\enc(\alpha_k) - \mathbf{x}_k||^2_2, 
    \end{align}
    where $||\nabla \func||_{\infty}$ represents the maximum absolute value of all elements in $\nabla \func(\cdot)$ over the set $\{\mathbf{x}_k\}^K_{k=1}$.
We note that $\sum_{k=1}^K||\enc(\alpha_k) - \mathbf{x}_k||^2_2$ is the training error of the encoding regression. Thus the end-to-end loss~\eqref{eq:main_formula}, depends on the training error of the encoding regression, multiplied by $||\nabla \func||_{\infty}$. 

{\bf Observation Two:} Unlike decoding regression, where the emphasis lies on generalization error, for encoding regression, the training error holds great significance and it is magnified by the factor of $||\nabla \func||_{\infty}$ in the end-to-end loss~\eqref{eq:main_formula}.
\end{itemize}

Considering observations one and two, we deduce that, in the input regression, both the training error and the smoothness of $\enc(\cdot)$ play significant roles. However, given that $||\nabla \func||_{\infty}$ tends to be large, especially in the case that the function is a deep neural network, prioritizing the reduction of the training error becomes more crucial.

\subsubsection{Proposed Regression Model}
As discussed, the generalization error of decoding regression depends on both its smoothness and the smoothness of encoding regression. We summarize these two smoothness factors as two distinct regularization terms, namely $\lambda_{\textrm{enc}}$ and $\lambda_{\textrm{dec}}$, which should be optimized jointly. Taking all into account, we propose the following pair of optimization statements to achieve encoding and decoding regression functions:
\begin{align}
\enc = \arg\min_{\mathbf{u} \in \mathbb{W}^2_2[a, b]} &\sum_{k=1}^{K} ||\mathbf{x}_k - \mathbf{u}(\alpha_k)||^2_2 \nonumber \\ &+ \lambda_{\textrm{enc}} \int^b_a ||\mathbf{u}^{\left({2}\right)}(t)||^2_2 \,dt  \label{eq:encoder_obj} \\
\dec = \arg\min_{\mathbf{u} \in \mathbb{W}^2_2[a, b]} &\sum_{j\in \mathcal{S}} ||\mathbf{\tilde{x}}_j - \mathbf{u}(\beta_j)||^2_2 \nonumber \\ &+ \lambda_{\textrm{dec}} \int^b_a ||\mathbf{u}^{\left({2}\right)}(t)||^2_2 \,dt \label{eq:decoding_obj}
\end{align}
where $a<b$ are arbitrary real numbers and $\mathbb{W}^2_2[a,b]$ containing all functions $g(\cdot)$ for which $g(\cdot)$ and $g'(\cdot)$ are absolutely continuous and $\int_{a}^{b} (g''(t))^2 \,dt < \infty$. Both \eqref{eq:encoder_obj} and \eqref{eq:decoding_obj} consist of two terms. The first term is the sum of squared errors at the initial points, and the second term measures the smoothness of the function up to the second derivative. The parameters $\lambda_{\textrm{dec}}$ and $\lambda_{\textrm{enc}}$ control the weight of the smoothing term in each optimization problem: As the $\lambda_{\textrm{enc}} \to 0$ ($\lambda_{\textrm{dec}} \to 0$), the estimation becomes closer to having an exact value at the input points. On the other hand, as $\lambda_{\textrm{enc}} \to \infty$ ($\lambda_{\textrm{dec}} \to \infty$), the objective returns the smoothest function with the minimum square error, resembling the best quadratic polynomial with the minimum square error.

Unlike Berrut schemes, which predominantly rely on (1) exact interpolation at input points and (2) numerically stable rational interpolation, we relax the exactness constraint. Instead, we put much more weight emphasis on the smoothness of the regression functions, as we have demonstrated its pivotal role in the end-to-end loss.

\subsubsection{Smoothing Spline}
The regression models proposed in \eqref{eq:encoder_obj} and \eqref{eq:decoding_obj} have been extensively studied in the literature. It can be demonstrated that the solution to the optimization problem is unique and equals to the {\bf natural cubic smoothing spline} with knots at $\{\alpha_k\}^K_{k=1}$ and $\{\beta_j\}_{j \in \mathcal{S}}$, respectively \cite{green1993nonparametric, wahba1975smoothing, de2001calculation}. In the limit where $\lambda_{\textrm{enc}} \to 0$ ($\lambda_{\textrm{dec}} \to 0$), the optimal function becomes a natural spline precisely passing through the data points. Conversely, as $\lambda_{\textrm{enc}} \to \infty$ ($\lambda_{\textrm{dec}} \to \infty$), the encoding (decoding) regression function converges to the best polynomial of degree three.

More precisely, given $n$ knots at $\{x_i\}^n_{i=1}$ and their corresponding function values $\{f(x_i)\}^n_{i=1}$, the smoothing spline model with smoothing parameter $\lambda$ estimates $\hat{f}(x) = \sum^n_{i=1} \theta_i \eta_i(x)$, where the set of functions $\{\eta_i(\cdot)\}^n_{i=1}$ are predefined basis functions, and $\{\theta_i\}^n_{i=1}$ are the model coefficients. The coefficients of the smoothing spline are computed from the following statement:
\begin{align}
    \hat{\mathbf{\theta}} = (\mathbf{M}^T\mathbf{M} + \lambda \mathbf{\Omega})^{-1}\mathbf{M}^T\mathbf{y},
\end{align}
where $\mathbf{y} = (f(x_1),\dots, f(x_n))$ and $\mathbf{M}, \mathbf{\Omega} \in \mathbb{R}^{n\times n}$ are defined as follow:
\begin{align}
    \mathbf{M}_{ij} = \eta_j(x_i) \quad \text{and} \quad \Omega_{ij} = \int^b_a \eta''_i(t)\eta''_j(t)\,dt
\end{align}
The regression function in the smoothing spline method is a linear model. Prediction on new points, as well as deriving the fitted coefficients, can be computed in $\mathcal{O}(K.d)$ operations in the encoding layer and $\mathcal{O}(|\mathcal{S}|.m)$ operations in the decoding layer using B-spline basis functions \cite{de1978practical,de2001calculation, hu1986complete, eilers1996flexible}. Consequently, the computational complexity of the proposed scheme is on par with Berrut rational interpolation \cite{jahani2022berrut}.

\section{Experimental Results}
In this section, we extensively evaluate the proposed scheme across various scenarios. Our assessments involve examining multiple deep neural networks as the prediction model and exploring the impact of different numbers of stragglers.

\subsection{Experimental Setup}
We evaluate the performance of the $\nprcc$ scheme in three different model architectures:
\begin{enumerate}
    \item \textbf{Shallow Neural Network}: We choose  LeNet5 \cite{lecun1998gradient} architecture that has two convolution layers and three fully connected layers, totaling $6\times 10^4$ parameters. We train the network on MNIST \cite{lecun2010mnist} dataset. The dataset consists of $32\times32$ grayscale images, and the network output is a ten-dimensional vector from the last softmax layer, determining the soft label associated with each class.

    \item \textbf{Deep Neural Network with low-dimensional output}: In this case, we evaluate the proposed scheme when the function is a deep neural network trained on color images in CIFAR-10 \cite{krizhevsky2009learning} dataset. We use the recently introduced RepVGG \cite{ding2021repvgg} network with around $26$ million parameters which was trained on the CIFAR-10\footnote{The pretrained weights can be found at \href{https://github.com/chenyaofo/pytorch-cifar-models}{https://github.com/chenyaofo/pytorch-cifar-models}}. The computation function in this scenario is $f:\mathbb{R}^{3072} \to \mathbb{R}^{10}$, where the input dimension is $32\times32\times3$, representing a random sample from the CIFAR-10 dataset, and the output is the class soft label predicted by the network.

    \item \textbf{Deep Neural Network with high-dimensional output}: Finally, we demonstrate the performance of the $\nprcc$ scheme in a scenario where the input and output of the computing function are high-dimensional, and the function is a relatively large neural network. We consider the Vision Transformer (ViT) \cite{dosovitskiy2020image} as one of the state-of-the-art base neural networks in computer vision for our prediction model, with more than $80$ million parameters (in the base version). The network was trained and fine-tuned on the ImageNet-1K dataset \cite{deng2009imagenet}\footnote{We use the official PyTorch pre-trained ViT network from \href{https://pytorch.org/vision/main/models/generated/torchvision.models.vit_b_16.html}{here}}.
\end{enumerate}

\setlength{\tabcolsep}{7pt}
\begin{table*}[t]
\caption{Comparison of different frameworks in terms of mean square error (MSE) and relative accuracy (RelAcc).}
\label{table:perf_mse_racc}
\vskip 0.15in
\begin{center}
\begin{small}
\begin{sc}
\begin{tabular}{lcc|cc|cc}
    \hline  & \\[-1.5ex]
    ~ 
    & \multicolumn{2}{c}{LeNet5} & \multicolumn{2}{c}{RepVGG}  & \multicolumn{2}{c}{ViT} \\
    $(N, K, |\mathcal{S}|)$
    &  \multicolumn{2}{c}{$(200, 50, 40)$}  & \multicolumn{2}{c}{$(200, 50, 80)$}  & \multicolumn{2}{c}{$(200, 50, 60)$} \\
    \cline{2-3}
    \cline{4-5}
    \cline{6-7}
     & \\[-1.5ex]
     \bf{Method} & \multicolumn{1}{c}{MSE} & \multicolumn{1}{c}{RelAcc} & \multicolumn{1}{c}{MSE} & \multicolumn{1}{c}{RelAcc} & \multicolumn{1}{c}{MSE} & \multicolumn{1}{c}{RelAcc} \\
    \hline  & \\[-1.5ex]
    \texttt{BACC}  & 1.06$\pm$ 0.31   & 0.99$\pm$ 0.01  & 2.09$\pm$ 0.21 & 0.87$\pm$ 0.03 & 0.99$\pm$ 0.156 & 0.80$\pm$ 0.07 \\
    \hline & \\[-1.5ex]
    $\nprcc^{Ag}$ & \textbf{0.83$\pm$ 0.27} &  \textbf{0.998$\pm$ 0.006} & 2.08$\pm$ 0.30 & 0.89$\pm$ 0.03 & 0.99$\pm$ 0.2 & \textbf{0.82$\pm$ 0.07}\\
    $\nprcc$    & \textbf{0.82$\pm$ 0.27}  &\textbf{0.998$\pm$ 0.006} & \textbf{1.87$\pm$ 0.17}& \textbf{0.90$\pm$ 0.02} & \textbf{0.84$\pm$ 0.12} & \textbf{0.82$\pm$ 0.06} \\
    \bottomrule
\end{tabular}
\end{sc}
\end{small}
\end{center}
\vskip -0.1in
\end{table*}

\begin{figure}[ht]
\vskip 0.2in
\begin{center}
\centerline{\includegraphics[width=\columnwidth]{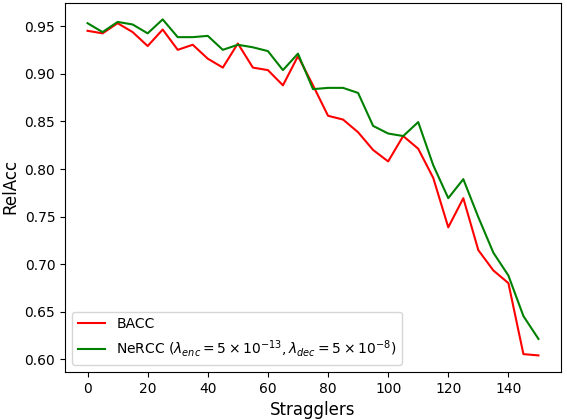}}
\caption{The relative accuracy of \texttt{BACC} and $\nprcc$ over a wide range of straggler numbers for $(N, K) = (200, 50)$ and RepVGG architecture.}
\label{fig:comp_berrut_racc_repvgg}
\end{center}
\vskip -0.3in
\end{figure}

\begin{figure}[ht]
\vskip 0.2in
\begin{center}
\centerline{\includegraphics[width=\columnwidth]{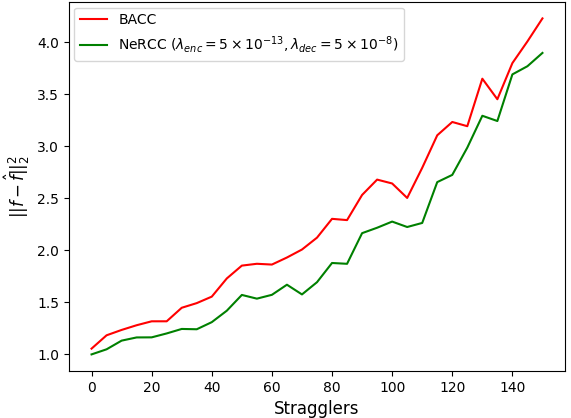}}
\caption{Reconstruction mean square error of \texttt{BACC} and $\nprcc$ across a wider range of straggler numbers for $(N, K) = (200, 50)$ and RepVGG architecture.}
\label{fig:comp_berrut_mse_repvgg}
\end{center}
\vskip -0.3in
\end{figure}

\textbf{Hyper-parameters:} The entire encoding and decoding process is the same for different functions, as we adhere to a non-parametric approach. The sole hyper-parameters involved are the two smoothing parameters ($\lambda_{\textrm{enc}}, \lambda_{\textrm{dec}}$).

\textbf{Interpolation Points:} We choose $\{\alpha_i\}^K_{k=1} = \cos\left(\frac{(2k-1)\pi}{2K}\right)$ and $\{\beta_n\}^N_{n=1} = \cos\left(\frac{(n-1)\pi}{N-1}\right)$ as Chebyshev points of the first and second kind, respectively, following the suggestions in \cite{jahani2022berrut}.

\textbf{Baseline:} We compare $\nprcc$ with the Berrut approximate coded computing (BACC) framework for distributed computing as a main baseline, introduced by \cite{jahani2022berrut}. The $\texttt{BACC}$ framework is used in \cite{jahani2022berrut} for training neural networks and in \cite{soleymani2022approxifer} for inference.

In addition, we evaluate the performance of the proposed scheme when both smoothing parameters are zero, $(\lambda_{\textrm{enc}}, \lambda_{\textrm{dec}}) = (0, 0)$, namely $\nprcc^{Ag}$, to be irrespective of the machine learning model. We call it $\nprcc^{Ag}$ because it makes the entire process the same for all machine learning models. Thus, it is a \textbf{model-agnostic} approach.

\textbf{Evaluation Metrics:} We use relative accuracy (RelAcc) and mean square error (MSE) as the evaluation metrics. The relative accuracy corresponds to the ratio of the base model prediction accuracy to the accuracy of the estimated model over initial data points $\{\mathbf{x}_k\}^K_{k=1}$. The  mean square error is equal to $E(||\func(\mathbf{x}) - \fhat(\mathbf{x})||^2_2) \approx \frac{1}{K}\sum^K_{k=1} ||\func(\mathbf{x}_k) - \fhat(\mathbf{x}_k)||^2_2$.

\begin{figure}[t]
\vskip 0.2in
\begin{center}
\centerline{\includegraphics[width=\columnwidth]{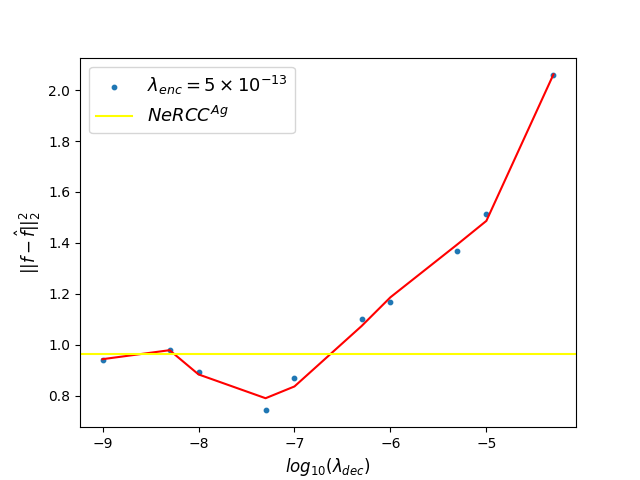}}
\caption{MSE of reconstruction for different values of $\lambda_{\textrm{dec}}$ in RepVGG architecutre trained on CIFAR10 dataset. The yellow line indicates the MSE of $\nprcc^{Ag}$ framework.}
\label{fig:dec_smooth}
\end{center}
\vskip -0.3in
\end{figure}

\begin{figure}[t]
\vskip 0.2in
\begin{center}
\centerline{\includegraphics[width=\columnwidth]{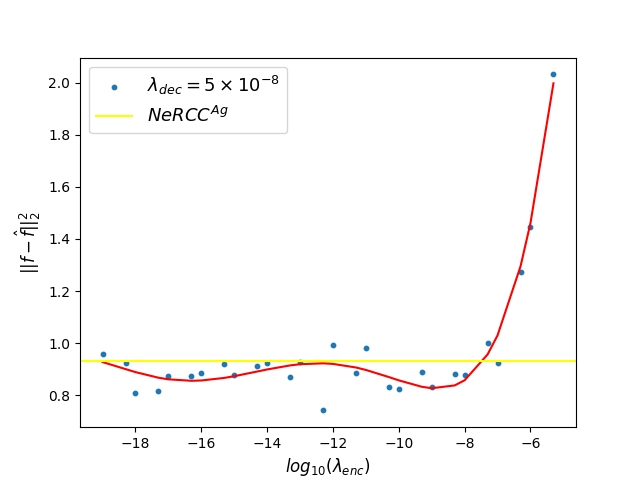}}
\caption{MSE of reconstruction for different values of $\lambda_{\textrm{enc}}$ in RepVGG architecutre trained on CIFAR10 dataset. The yellow line indicates the MSE of $\nprcc^{Ag}$ framework.}
\label{fig:enc_smooth}
\end{center}
\vskip -0.3in
\end{figure}

\subsection{Performance Evaluation}
 Table~\ref{table:perf_mse_racc} presents a comparison between \texttt{BACC} and our proposed \texttt{NeRCC} scheme. Presenting both metrics side by side provides a comprehensive view of the performance evaluation. Table~\ref{table:perf_mse_racc} shows that MSE of $\nprcc$ is significantly less than \texttt{BACC} scheme in all architectures. In addition, the RelAcc of $\nprcc$ is comparable to \texttt{BACC}. This is mainly because accuracy is computed from quantized function outputs. the MSE metric reveals a notable improvement in $\nprcc$. Table~\ref{table:perf_mse_racc} indicates an average $22\%$, $11\%$, and $15\%$ improvement in mean square error for LeNet, RepVGG, and ViT architectures, respectively.

In the subsequent analysis, we evaluate the performance of $\nprcc$ compared to \texttt{BACC} for a diverse range of stragglers. For a constant number of stragglers, we execute \texttt{BACC} and $\nprcc$ approaches with the same set of input data points multiple times and record the average RelAcc and average MSE metrics. Figures~\ref{fig:comp_berrut_racc_repvgg} and \ref{fig:comp_berrut_mse_repvgg} present the performance of both schemes on the RepVGG architecture for RelAcc and MSE metrics, respectively (refer to Figure~\ref{fig:comp_berrut_all} in Appendix \ref{sec:apx_perf} for complete results). As illustrated in Figure~\ref{fig:comp_berrut_all}, the proposed scheme consistently outperforms \texttt{BACC} across nearly all values for the number of stragglers,  emphasizing the significance of regression model smoothness.

\subsection{Smoothness effect}\label{sec:smooth_eff}
We further investigate the effect of $\nprcc$ \emph{smoothness parameters} on the final results. By fixing $\lambda_{\textrm{enc}}$ ($\lambda_{\textrm{dec}}$), we examine how the MSE changes as $\lambda_{\textrm{dec}}$ ($\lambda_{\textrm{enc}}$) varies. Figures~\ref{fig:enc_smooth} and \ref{fig:dec_smooth} depict the results for the RepVGG architecture and the configuration $(N, K, |\mathcal{S}|) = (200, 50, 20)$. The yellow line shows the MSE of $\nprcc^{Ag}$. As depicted in both figures, increasing the smoothing parameter from zero initially reduces the overall mean square error, followed by a rapid exponential increase. This indicates that there is an optimal non-zero smoothness level


Furthermore, we analyze the performance of $\nprcc^{Ag}$ against $\nprcc$ and \texttt{BACC}.  The results are presented in Figure~\ref{fig:comp_ns} in Appendix \ref{sec:apx_smoothing_effect}. $\nprcc$ hyper-parameters are optimized for 40, 50, and 60 stragglers in the LeNet, RepVGG, and ViT architectures, respectively. Next, we compare the performance of those tuned $\nprcc$ with $\nprcc^{Ag}$ and \texttt{BACC} for a wide range of straggler numbers. As depicted in Figure~\ref{fig:comp_ns}, the MSE of $\nprcc$ outperforms both $\nprcc^{Ag}$ and \texttt{BACC} in higher straggler regimes. However, in low straggler regimes, where there is sufficient data redundancy in our coded samples (the values of $\{\beta_n\}^N_{n=1}$ and $\{\alpha_k\}^K_{k=1}$ ensure a satisfactory number of $\beta_n$ around each $\alpha_k$), $\nprcc^{Ag}$ has both smoothness and accuracy advantages over the other two approaches. We noted that the $\nprcc^{Ag}$ outperforms \texttt{BACC} in all model architectures and for different numbers of stragglers. Recall that $\nprcc^{Ag}$ is a model-agnostic approach.




\subsection{Coded points}
We also compare the coded points $\{\mathbf{\tilde{x}_n}\}_{n=1}^N$ sent to the workers in $\nprcc$ and \texttt{BACC} schemes. The results, shown in Figure~\ref{fig:coded_samples_all} in Appendix~\ref{sec:apx_coded_sample}, demonstrate that \texttt{BACC} coded samples exhibit high-frequency noise which causes the scheme to approximate original prediction worse than $\nprcc$.

\section{Related Works}
Mitigating stragglers or slow workers is one of the main challenges of distributed computing systems, such as prediction serving systems, and has been extensively studied in recent years. Traditional approaches rely on repetition, as seen in works such as \cite{zaharia2008improving, dean2013tail}, where a worker's task is reassigned to another worker only if its progress rate in completing that task falls below a certain threshold. In recent years, coding-based techniques, termed coded computing, have emerged as a promising approach to enhancing the efficiency of distributed computing systems. These methods offer significant improvements in latency and overhead by mitigating the impact of stragglers \cite{lee2017speeding, kosaian2018learning, yu2019lagrange, narra2019distributed, wang2019erasurehead, wang2019fundamental}. In the coded computing approach, each node is given a combination of inputs, rather than the raw data, to process. The objective is to design the code in such a way that the results can be recovered from any set of available results on coded data while ignoring the results of stragglers.
 Initially, coded computing approaches were designed to exploit the structure of computations, limiting their applications to highly structured tasks like matrix multiplication and polynomial computation. The other disadvantage is that the required number of workers becomes prohibitively large.
Some effort has been made to extend those ideas to include non-polynomials by approximating the function with polynomials~\cite{so2020scalable}. Still, the scope of this approach is very limited. Subsequently, learning-based methods have been introduced to reduce the number of necessary workers required to mitigate stragglers \cite{kosaian2019parity, kosaian2020learning, dinh2021coded}. However, these models need to train a specific neural network for each computing function. In addition, that approach offers resistance to only a very small number of stragglers. More recently, a numerically stable, model-agnostic, and straggler-resilient method based on Berrut rational interpolation and approximate estimation of the original function at the input data points was introduced by \cite{jahani2022berrut}. In the scheme of~\cite{jahani2022berrut} the main effort was to choose a pole-free rational function that exactly interpolation at initial data points to have exact and numerically stable interpolation. The approach employed in this paper is based on assessing the end-to-end performance to derive design criteria that place value on the smoothness of the decoding and encoding functions in an intriguing manner.

\section{Conclusion}
In this paper, we have introduced $\nprcc$, a general and resilient coded computing framework for prediction serving systems. Our framework consists of two regression layers and a central computing layer, sandwiched between the two regression layers. In contrast to all existing coded computing schemes, where the primary focus is on exact interpolation, we prioritize the final accuracy of recovering original predictions as our main objective. We have stated that the overall objective reveals an underlying interplay between encoding and decoding regression, and we have proposed a technique to solve the mentioned coupled regression problem. Our experiments on MNIST, CIFAR-10, and ImageNet-1K datasets, using various model architectures such as LeNet, RepVGG, and ViT, show that $\nprcc$ and its model-agnostic version ($\nprcc^{Ag}$) outperform the state-of-the-art (\texttt{BACC}) in a wide range of straggler numbers, as well as different metrics, by up to 23\%.



\bibliographystyle{icml2024}

\newpage
\appendix
\onecolumn
\section{Appendix}
\subsection{Performance comparison}\label{sec:apx_perf}

\begin{figure}[ht]
\vskip 0.2in
\centering     
\subfigure[$(N, K)=(100, 25)$, ViT model]{\label{fig:comp_all_mse_vit}\includegraphics[width=0.33\textwidth]{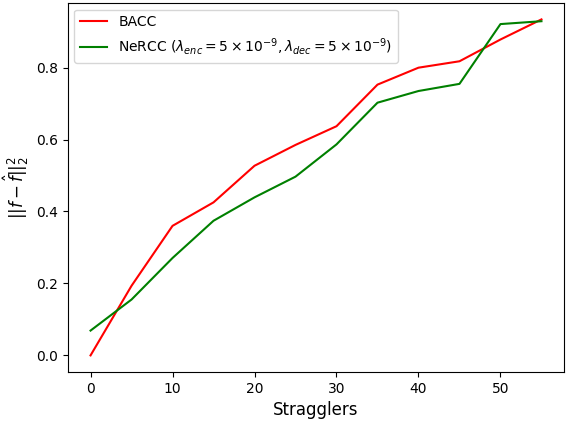}}
\subfigure[$(N, K)=(200, 50)$, RepVGG model]{\label{fig:comp_all_mse_cifar}\includegraphics[width=0.33\textwidth]{figs/mse_cifar_comp_berrut.png}}
\subfigure[$(N, K)=(200, 50)$, LeNet model]{\label{fig:comp_all_mse_mnist}\includegraphics[width=0.33\textwidth]{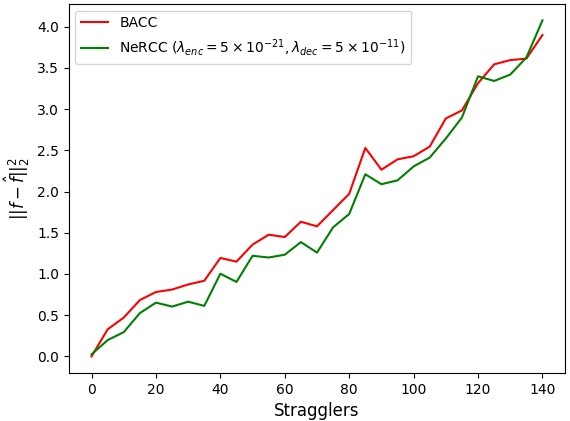}}
\\
\subfigure[$(N, K)=(100, 25)$, ViT model]{\label{fig:comp_all_racc_vit}\includegraphics[width=0.33\textwidth]{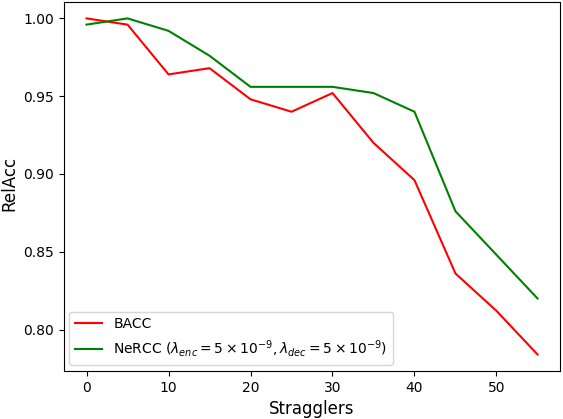}}
\subfigure[$(N, K)=(200, 50)$, RepVGG model]{\label{fig:comp_all_racc_cifar}\includegraphics[width=0.33\textwidth]{figs/racc_cifar_comp_berrut.png}}
\subfigure[$(N, K)=(200, 50)$, LeNet model]{\label{fig:comp_all_racc_mnist}\includegraphics[width=0.33\textwidth]{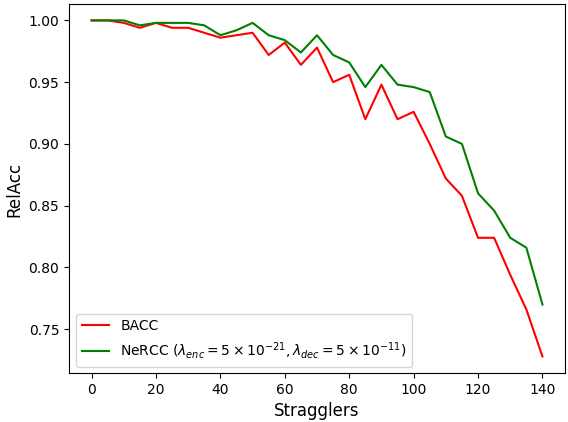}}

\caption{Performance comparison of straggler resistance between \texttt{BACC} and $\nprcc$ frameworks across all scenarios and metrics. The difference for MSE as a metric is illustrated in \cref{fig:comp_all_mse_vit,fig:comp_all_mse_cifar,fig:comp_all_mse_mnist}, while the comparison of the relative accuracy is depicted in \cref{fig:comp_all_racc_vit,fig:comp_all_racc_cifar,fig:comp_all_racc_mnist}.}
\label{fig:comp_berrut_all}
\vskip -0.2in
\end{figure}

As depicted in Figure~\ref{fig:comp_berrut_all}, the $\nprcc$ scheme consistently outperforms the \texttt{BACC} scheme across a wide range of straggler numbers, for different model architecture, and in terms of both MSE and RelAcc metrics.

\subsection{Smoothing Effect}\label{sec:apx_smoothing_effect}
\begin{figure}[h]
\vskip 0.2in
\centering     
\subfigure[$(N, K)=(200, 50)$, LeNet model]{\label{fig:comp_ns_mse_mnist}\includegraphics[width=0.33\textwidth]{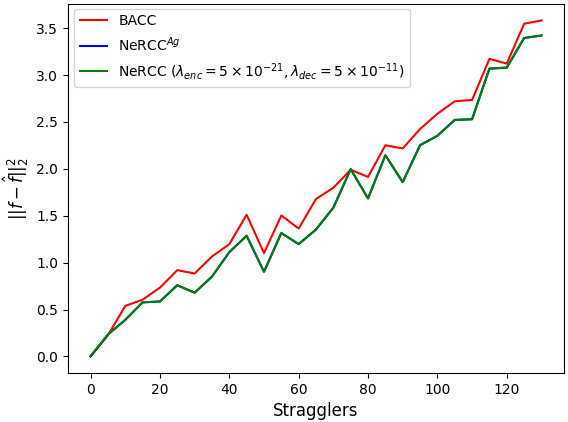}}
\subfigure[$(N, K)=(200, 50)$, RepVGG model]{\label{fig:comp_ns_mse_cifar}\includegraphics[width=0.33\textwidth]{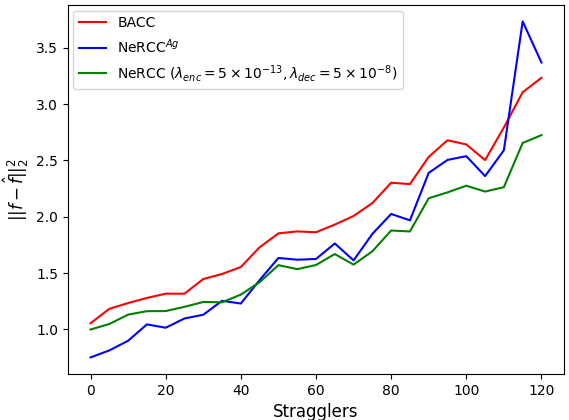}}
\subfigure[$(N, K)=(100, 25)$, ViT model]{\label{fig:comp_ns_mse_vit}\includegraphics[width=0.33\textwidth]{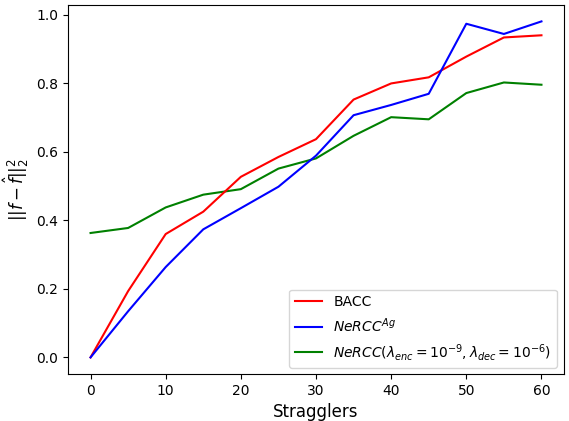}}
\caption{Comparison of Mean square error for $\nprcc^{Ag}$ and $\nprcc, \texttt{BACC}$ over different number of stragglers for \ref{fig:comp_ns_mse_mnist} LeNet, \ref{fig:comp_ns_mse_cifar} RepVGG, and \ref{fig:comp_ns_mse_vit} ViT networks as the machine learning models.}
\label{fig:comp_ns}
\vskip -0.2in
\end{figure}

Figure~\ref{fig:comp_ns} illustrates the impact of using smoothing parameters in $\nprcc$ and presents the overall performance comparison between $\nprcc$ and its model-agnostic version, $\nprcc^{Ag}$, along with the \texttt{BACC} framework. It is evident that for less complex functions like LeNet, both $\nprcc$ and $\nprcc^{Ag}$ yield the same results. However, for more complex functions such as ViT and RepVGG, while $\nprcc^{Ag}$ performs better in scenarios with a lower number of stragglers (benefiting from data point redundancy), it notably falls behind $\nprcc$ in higher straggler regimes.

\subsection{Coded samples}\label{sec:apx_coded_sample}
\begin{figure}[h]
\vskip 0.2in
\centering     
\subfigure[Initial inputs]{\label{fig:coded_samples_a}\includegraphics[scale=0.25]{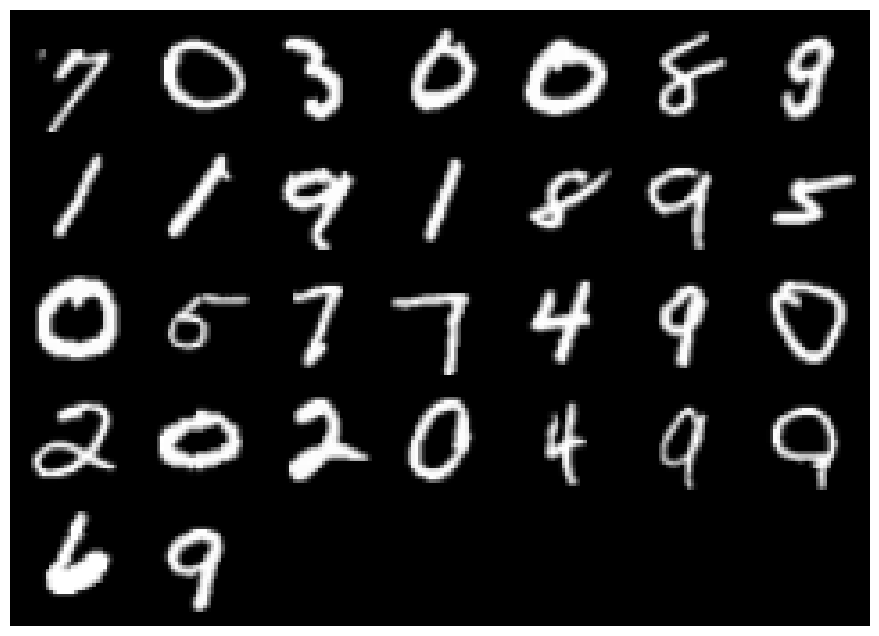}}
\\
\subfigure[\texttt{BACC} coded samples]{\label{fig:coded_samples_b}\includegraphics[width=0.35\textwidth]{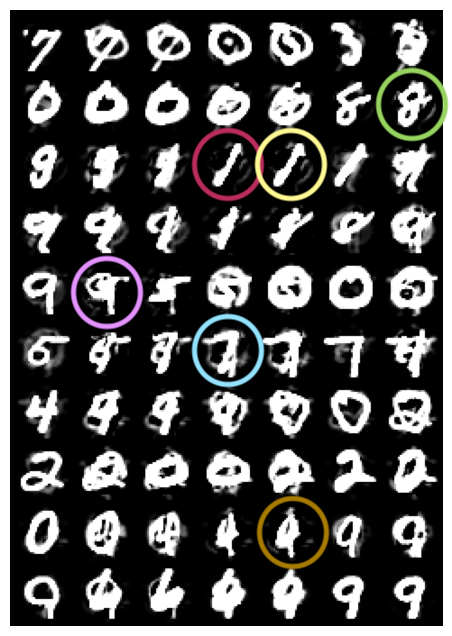}}
\subfigure[$\nprcc$ coded samples]{\label{fig:coded_samples_c}\includegraphics[width=0.35\textwidth]{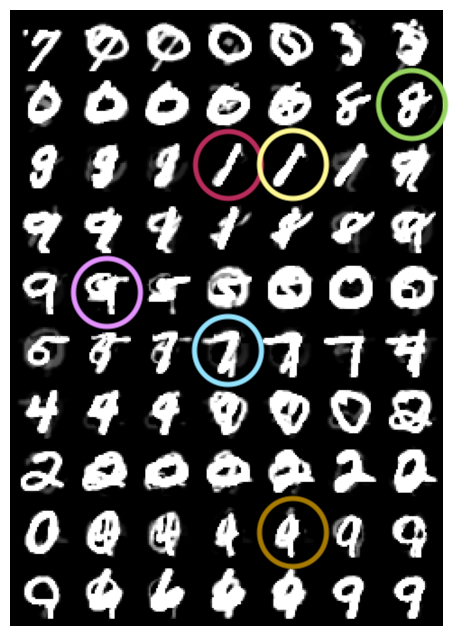}}
\caption{Comparison of coded samples between \texttt{BACC} and $\nprcc$ frameworks. Figure~\ref{fig:coded_samples_a} represents the initial data points $\{\mathbf{x}_k\}^K_{k=1}$ for $K=30$. Figures~\ref{fig:coded_samples_b} and \ref{fig:coded_samples_c} display $N=70$ coded samples $\{\mathbf{\tilde{x}}_n\}^N_{n=1}$ from \texttt{BACC} and $\nprcc$, respectively. Samples with clear differences are highlighted with the same color.}
\label{fig:coded_samples_all}
\vskip -0.2in
\end{figure}

Figures~\ref{fig:coded_samples_b} and \ref{fig:coded_samples_c} display coded samples generated by \texttt{BACC} and $\nprcc$, respectively, derived from the same initial data points depicted in Figure~\ref{fig:coded_samples_a}. These samples are presented for the MNIST dataset with parameters $(N, K) = (70, 30)$. From the figures, it is apparent (Specifically in paired ones that are shown with the same color) that while both schemes' coded samples are a weighted combination of multiple initial samples, \texttt{BACC}'s coded samples exhibit high-frequency noise. This observation suggests that $\nprcc$ regression functions produce more refined coded samples without any disruptive noise.

\end{document}